\documentclass[letterpaper, 10 pt, journal, twoside]{ieeetran}   %
\IEEEoverridecommandlockouts                              %

\usepackage{enumerate}
\usepackage{graphics} %
\usepackage{epsfig} %
\usepackage{amsmath} %
\usepackage{amssymb}  %
\usepackage{array}
\usepackage[font=footnotesize]{caption}
\usepackage{subcaption}
\usepackage[tight]{units}
\usepackage{glossaries}
	\newacronym{hyq}{HyQ}{Hydraulically actuated Quadruped}
	\newacronym{lf}{LF}{Left-Front}
	\newacronym{rf}{RF}{Right-Front}
	\newacronym{lh}{LH}{Left-Hind}
	\newacronym{rh}{RH}{Right-Hind}
	\newacronym{imu}{IMU}{Inertial Measurement Unit}
	\newacronym{dofs}{DoFs}{Degrees of Freedom}
	\newacronym{com}{CoM}{Center of Mass}
	\newacronym{cop}{CoP}{Center of Pressure}
	\newacronym{zmp}{ZMP}{Zero Moment Point}
    \newacronym{cwc}{CWC}{Contact Wrench Cone}
    \newacronym{awp}{AWP}{Actuation Wrench Polytope}
    \newacronym{fws}{FWP}{Feasible Wrench Polytope}
	\newacronym{grfs}{GRFs}{Ground Reaction Forces}
	\newacronym{cmm}{CMM}{Centroidal Momentum Matrix}
	\newacronym{rnea}{RNEA}{Recursive Newton-Euler Algorithm}
	\newacronym{eom}{EoM}{Equation of Motions}
	\newacronym{to}{TO}{Trajectory Optimization}
    \newacronym{qp}{QP}{Quadratic Programming}
	\newacronym{micp}{MICP}{Mixed-Integer Convex Programming}
	\newacronym{miqcqp}{MIQCQP}{Mixed-Integer Quadratically Constrained
    Quadratic Programming}
\newcommand{\mat}[1]{\ensuremath{\begin{bmatrix}#1\end{bmatrix}}}
\DeclareMathOperator*{\st}{s.t.}
\newcommand{\sref}[1]{Section~\ref{#1}}
\newcommand{\fref}[1]{Fig.~\ref{#1}}

\newcommand{\tref}[1]{Table~\ref{#1}}

\title{Simultaneous Contact, Gait and Motion Planning for Robust
Multi-Legged Locomotion via Mixed-Integer Convex Optimization}

\author{Bernardo Aceituno-Cabezas$^{1}$, Carlos Mastalli$^{2}$,
Hongkai Dai$^{3}$, Michele Focchi$^{2}$, Andreea Radulescu$^{2}$,
\\ Darwin G. Caldwell$^{2}$, Jos\'e Cappelletto$^{1}$, Juan C. Grieco$^{1}$,
Gerardo Fern{\'a}ndez-L{\'o}pez$^{1}$, and Claudio Semini$^{2}$%
\thanks{Manuscript received: September, 10, 2017; Accepted November, 16, 2017. This paper was recommended for publication by Editor Paolo Rocco upon evaluation of the Associate Editor and Reviewers' comments.}
\thanks{The authors are with (1) the Mechatronics Research Group, Sim{\'o}n
Bol{\'i}var University {\tt \small \{12-10764, cappelletto, jcgrieco,
gfernandez\}@usb.ve}, (2) are with the Department of Advanced Robotics,
Istituto Italiano di Tecnologia {\tt \small \{firstname. surname\}}@iit.com,
and (3) is with the Toyota Research Institute {\tt \small hongkai.dai@tri.global}}%
\thanks{Digital Object Identifier (DOI): see top of this page.}
}

\begin{document}

\maketitle

\begin{abstract}
Traditional motion planning approaches for multi-legged locomotion 
divide the problem into 
several stages, such as contact search and trajectory generation.  However,
reasoning about contacts and motions simultaneously is crucial for the
generation of complex whole-body behaviors. Currently, coupling theses problems has
required either the assumption of a  fixed gait sequence and flat 
terrain condition, or non-convex
optimization with intractable computation time. In this paper, we propose a
mixed-integer convex formulation to plan simultaneously contact locations,
gait transitions and motion, in a computationally efficient fashion. In
contrast to previous works, our approach is not limited to flat terrain
nor to a pre-specified gait sequence. Instead, we incorporate the friction
cone stability margin, approximate the robot's torque limits, and plan the gait using 
mixed-integer convex constraints. We experimentally validated 
our approach on the HyQ robot 
by traversing different challenging terrains, 
where non-convexity and flat terrain assumptions might lead
to sub-optimal or unstable plans. 
Our method increases the motion robustness while
keeping a low computation time.
\end{abstract}

\section{Introduction}

\IEEEPARstart{P}{lanning} motions for multi-legged robots can be a challenging task, as it involves 
both the discrete choice of the gait sequence, and continuous decisions on foot 
location and robot dynamics. Because of this, most traditional approaches 
\cite{deits2014footstep,winkler2015planning,aceituno2017convex,tonneau2016efficient} 
simplify the planning problem by decoupling it into two stages: 1) find a sequence of
contact locations; and 2) generate body trajectories based on a stability
metric, like the \gls{zmp} \cite{vukobratovic1969zmp} or the generalized \gls{cwc}
\cite{hirukawa2006universal,wieber2002stability} stability criterion. 
Ignoring the dynamics in the first stage might restrict the number of feasible
motion on the second one, when dynamics are considered. Hence, 
in this paper we combine these two stages 
into one single problem, in order to efficiently generate complex behaviors.

\gls{to} has emerged as a solution to the coupled motion planning problem 
\cite{mastalli2017trajectory,winkler17b,dai2014whole,Posa14,neunert2017trajectory}. 
Unfortunately, even state of 
the art approaches are often restricted to simple environments, a fixed gait, or require 
intractable computation time. Furthermore, none of them can guarantee global optimality, 
and can easily fall into local minima, since these are posed as non-convex
optimization problems. On the other hand, \gls{micp} proved to be an efficient tool to solve
multi-contact motion planning problems. In \cite{deits2014footstep,aceituno2017convex}, 
the authors first find a set of safe surfaces, and then use integer variables to assign each
contact to one surface. Further works incorporate a convex or mixed-integer convex dynamic
model in the formulation \cite{ponton2016convex,valenzuela2016mixed}, which can efficiently
optimize contacts and motions. Unfortunately, these approaches do not incorporate torque
constraints nor maximize any generalized stability margin. Moreover, they do not reason
about the sequence of contacts, being limited to a fixed gait sequence.

\begin{figure}[t]
  \centering
  \includegraphics[width=0.87\linewidth]{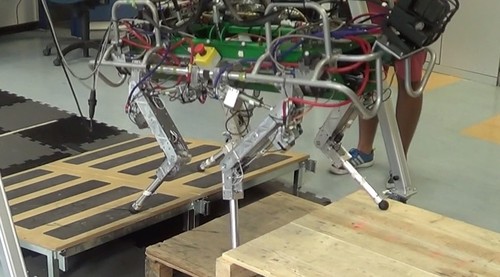}
  \caption{The proposed approach generates robust plans in 
  challenging terrain, simultaneously reasoning about contacts and motions. 
  This figure shows \acrshort{hyq} executing an optimized motion with 
  non-coplanar contacts.}
  \label{fig:figintro}
  \vspace{-19pt}
\end{figure}

In this work, we introduce a novel \gls{micp} formulation for non-gaited multi-legged 
locomotion on challenging terrain. Here, we formulate a single mixed-integer convex 
optimization problem that simultaneously plans contacts and motions. In contrast to other
approaches, our method guarantees dynamic feasibility and optimizes gaits sequences in
challenging terrains. For this, we incorporate convex representations of friction cones,
torque limits, and gait planning constraints. For the friction cone constraints, we
exploit the segmentation of the terrain and pose the stability margin computation as
a convex constraint. This optimization problem can then be solved efficiently to its
\textit{global optimum},
using off-the-shelf optimization solvers. We validated 
our approach by traversing different
challenging terrains with the \gls{hyq} robot, shown in Fig. \ref{fig:figintro}. Our approach
generates robust motions, even during gait transitions and non-coplanar contact scenarios.

The rest of this paper is organized as follows: Section II presents our simultaneous 
contact and motion planning approach. Section III presents our whole-body controller. 
Section IV presents a set of experiments on \gls{hyq} traversing challenging terrains, and
Section V discusses and concludes on the contributions of this work.
 
\section{Simultaneous Contact and Motion Planning}

In this section, we describe our formulation of the simultaneous contact and motion 
planning problem using Mixed-Integer Convex Optimization.

\begin{figure}[t]
  \centering
  \includegraphics[height=0.5\linewidth]{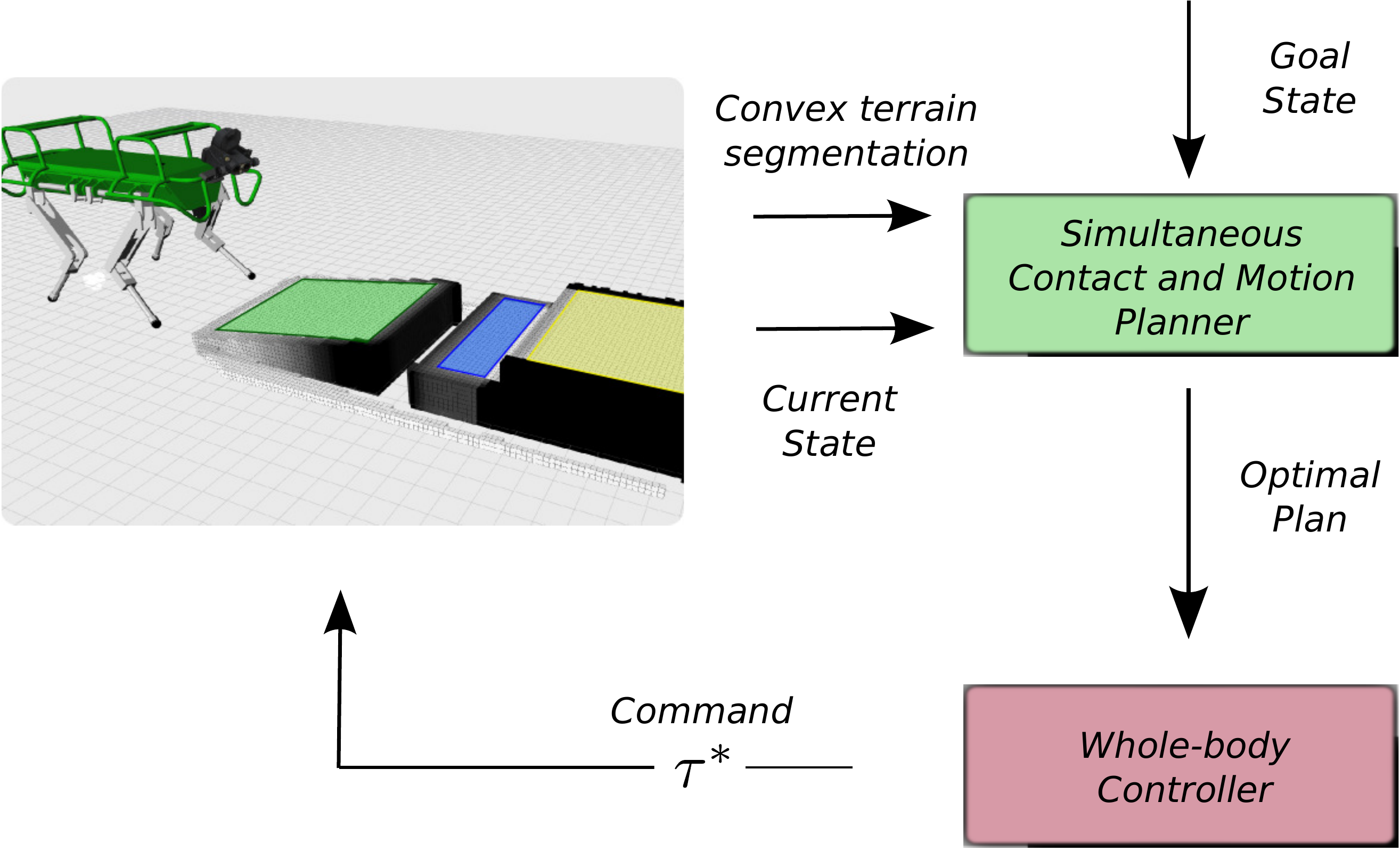}
  \caption{Overview of our convex trajectory optimization approach. Here, we compute 
  contact locations and motions within a single mixed-integer convex program, given a goal 
  state and a convex segmentation of the safe terrain.}
  \label{fig:figoverview}
  \vspace{-12pt}
\end{figure}

\subsection{Approach Overview}
	
Let us consider a robot with $n_l$ legs and a locomotion plan, discretized on $N$ time-knots. 
We formulate a \gls{micp} to find the optimal contact locations and motions, given a goal
position $\mathbf{r}_G$ and a convex segmentation of the terrain. An illustration of
this approach is shown in Fig. \ref{fig:figoverview}.

We describe the dynamic evolution of the system with a centroidal model (\sref{sec:dyn}). 
We include gait planning as part
of the optimization (\sref{sec:gait}), describing the motion through time-slots over which legs
swing. This evolution is 
subject to robot and environmental constraints such as: approximated kinematic 
reachability (\sref{sec:contact}), friction cone constraints (\sref{sec:frict}),
and approximated torque limits (\sref{sec:torq}). 
describing the motion through time-slots over which we will generate
leg swings. We adopt the formulation in \cite{ponton2016convex} 
to further address the non-linearities of the angular dynamics by relying
on a convex decomposition of bilinear terms, making our optimization problem convex
(\sref{sec:conv_ang_momentum_model}). \vskip6pt

\subsection{Centroidal Dynamics}\label{sec:dyn}

\begin{figure}[t]
  \centering
  \includegraphics[height=0.5\linewidth]{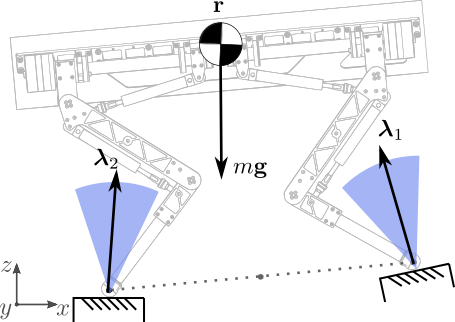}
  \caption{Centroidal dynamic model of a quadruped.}
  \label{fig:figcent}
  \vspace{-6pt}
\end{figure}

We describe the evolution of the system using a centroidal 
dynamics model \cite{orin2013centroidal,daiplanning}, as depicted in Fig. 
\ref{fig:figcent}. In this case, the dynamics of the system are written as:

\begin{equation}
\label{dyn}
\begin{bmatrix}
m \ddot{\mathbf{r}} \\
\dot{\mathbf{k}}
\end{bmatrix} = \begin{bmatrix}
m \mathbf{g} + \sum_{l = 1}^{n_l} \boldsymbol{\lambda}_{l} \\
\sum_{l = 1}^{n_l} (\mathbf{p}_{l} - \mathbf{r}) \times \boldsymbol{\lambda}_{l}
\end{bmatrix},
\end{equation}
where variables are named as specified in Table \ref{table:model}, $m$ is the mass of the 
robot, and $\textbf{g}$ is the gravity vector. This formulation of the dynamics is 
entirely \textit{convex}, except for the cross product to compute the torque at center of mass. \vskip6pt

\textit{Remark 1:} Note that \eqref{dyn} neglects the 
effect of contact torques. This is done because most multi-legged 
platforms have approximate point contacts, 
without support surface to generate moments. \vskip6pt

Furthermore, the dynamics of the model are discretized in knots of $\Delta t$ seconds, 
over which we will perform backward Euler integration.

\begin{table}[t]
	\caption{Variables of the trajectory optimization}
	\begin{tabular}{l l}
		$\mathbf{r} = [r_x, r_y, r_z]^T$ & \gls{com} in the world frame\\
        $\mathbf{k}  = [k_{x}, k_{y}, k_{z}]^T$ & Centroidal angular momentum \\
        $\mathbf{p}_l  = [p_{l_x}, p_{l_y}, p_{l_z}]^T$ & End-effector positions in the 
        world frame \\
        $\boldsymbol{\lambda}_l  = [\lambda_{l_x}, \lambda_{l_y}, \lambda_{l_z}]^T$ & 
        Contact force at the $l^{th}$ end-effector \\
        $N_t, N_k$ & Number of time-slots and knots per swing\\
        $N_f, N_r$ & Number of contacts and convex regions \\
	\end{tabular}
    \label{table:model}
      \vspace{-15pt}
\end{table}

\begin{figure*}[t]
  \begin{minipage}{.5\textwidth}
      \centering   \includegraphics[height=0.5\linewidth]{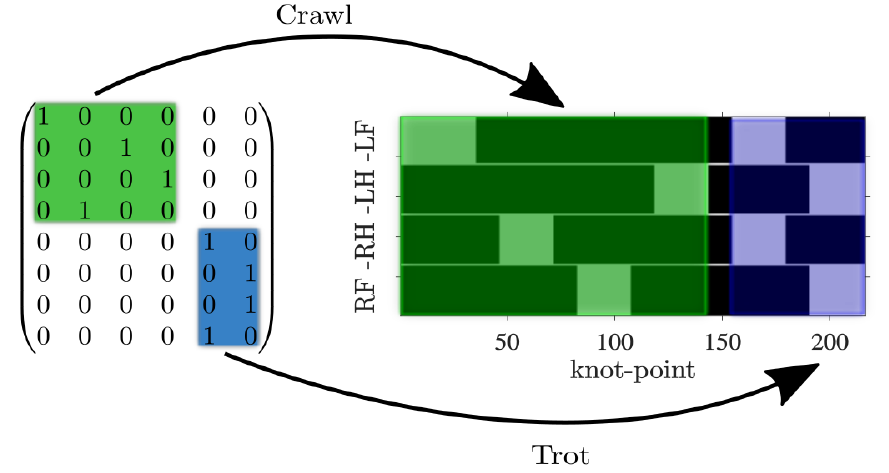}
      \label{fig:sub:figgait}
      \captionof{figure}{Gait transition matrix and equivalent phase diagram.
      R, L,\\ H and F represent right, left, hind and front supports.}
      \vspace{-15pt}
  \end{minipage}
  \hspace{1pt}
  \begin{minipage}{.5\textwidth}
    \centering
  \includegraphics[height=0.5\linewidth]{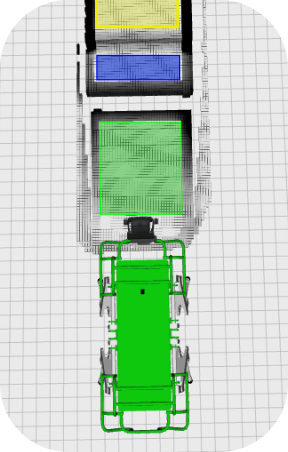}   
  \hspace{3pt}
  \vrule
  \hspace{3pt}
  \includegraphics[height=0.5\linewidth]{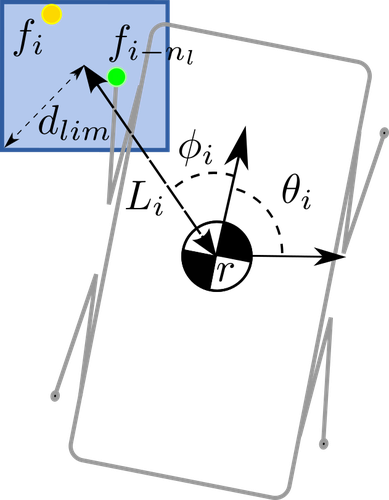}
    \captionof{figure}{\textit{Left:} Convex segmentation of safe contact surfaces. 
    \textit{Right: }Approximate kinematic reachability constraints.}
    \label{fig:figkin}
    \vspace{-15pt}
  \end{minipage}
  \label{fig:f1}
\end{figure*}

\subsection{Gait Sequence}\label{sec:gait}

In order to plan gait transitions, we represent the motion through 
$N_f$ contact locations and $N_t$ time-slots over which the legs will swing between adjacent contacts
, and introduce a binary \textit{transition matrix} $\mathbf{T} \in \{0,1\}^{N_f \times N_t}$. 
Here, $\mathbf{T}_{i j} = 1$ means that the robot will move to the $i^{th}$ contact location at the 
$j^{th}$ time-slot, as shown in Fig. 4. Since each contact location in the plan is 
reached once, we enforce that:
\begin{equation}
\sum_{j = 1}^{N_t} \mathbf{T}_{i j} = 1 \ , \ \forall i = 1,..,N_f.
\end{equation}
Additionally, we enforce that each cycle of $n_l$ contacts must be reached before the 
next transfer cycle starts. For this, we define a vector $ \mathbf{t} = \mathbf{T}
\begin{bmatrix}
1 & \hdots & N_t
\end{bmatrix}^T $ which computes the corresponding slot assigned to each movement. 
Furthermore, since the gait follows a sequentially ordered contact plan, we enforce the 
following constraint:
\begin{equation}
\mathbf{t}_j > \mathbf{t}_{j-n_{l}} \ , \ \forall j = n_{l} + 1, \dots, N_t.
\end{equation}	
Terrain heuristics can also be added in order to guide the gait through 
different environments. These can be incorporated in the formulation as shown in 
\cite{aceituno2017convex}.

\subsection{Contact Location}\label{sec:contact}

In order to simplify the formulation, we only optimize the contact locations and 
contact timing, leaving the end-effector trajectory as an interpolation between adjacent 
contacts in the plan. Since swing phases do not contribute to the centroidal dynamics, as force in 
the leg becomes null, this simplification has no effect on the resulting plan.

As part of the decision variables, we describe contact locations for $n_l$ end-effectors
using an array of $N_f$ vectors in $\mathbb{R}^4$, ordered by end-effector number,
of the form:
\begin{equation}\nonumber
\mathbf{f} = (f_x, f_y, f_z, \theta),
\end{equation}
representing the position of each contact in Cartesian space and the yaw orientation of 
the trunk when transitioning to that contact, neglecting roll and pitch positions.\vskip 6pt

\subsubsection{Safe-region assignment}\label{sec:reg}
Here, we will invert the problem of avoiding obstacles by constraining the contacts to lie
within one of $N_r$ \textit{convex safe contact surfaces}, shown colored in \fref{fig:figkin} (\textit{left}). Each surface is represented as a polygon
$\mathcal{R} = \{\mathbf{c} \in \mathbb{R}^3 | A_\mathrm{r} \mathbf{c} \leq b_\mathrm{r} \}$. 
The assignment of contacts to these surfaces is done through a binary decision matrix
$\mathbf{H} \in \{ 0,1 \}^{N_f\times N_r}$. The constraints, for the $i^{th}$ contact, are:
\begin{equation}
\sum_{\mathrm{r} = 1}^{N_r}{\mathbf{H}_{i \mathrm{r}}} = 1,
\end{equation}
\vspace{-6pt}
\begin{equation}
\mathbf{H}_{i \mathrm{r}} \Rightarrow A_\mathrm{r} \mathbf{f}_i \leq b_r,
\end{equation}
where the $\Rightarrow$ (implies) operator is represented with big-M formulation 
\cite{richards2005mixed}. Such surfaces can be easily obtained with segmentation algorithms. 
\vskip 6pt

\begin{figure*}[t]
    \begin{minipage}{.5\textwidth}
		\centering
    	\includegraphics[height=0.4\linewidth]{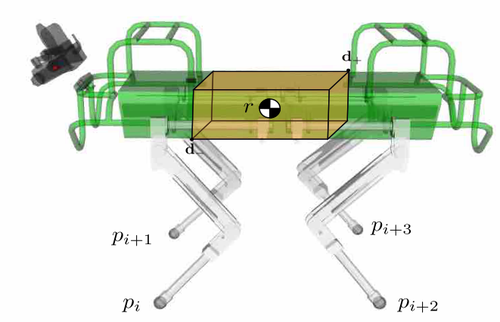}
    	\label{fig:figkin2}
        \captionof{figure}{We incorporate bounding box constraints, shown in 
        orange,\\ to ensure the \gls{com} remains in its workspace.}
      	\vspace{-12pt}
  	\end{minipage}
  	\begin{minipage}{.5\textwidth}
    	\centering
    	\includegraphics[height=0.4\linewidth]{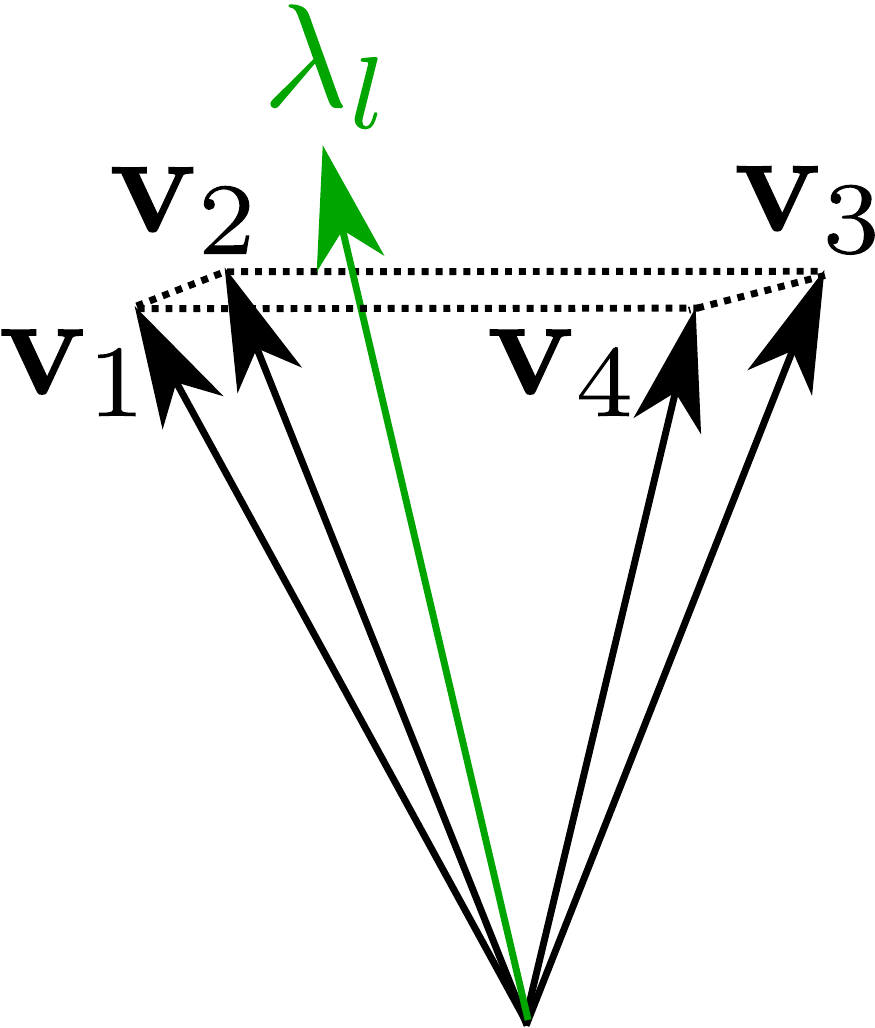}
        \hspace{3pt}
        \vrule
        \hspace{3pt}
        \includegraphics[height=0.4\linewidth]{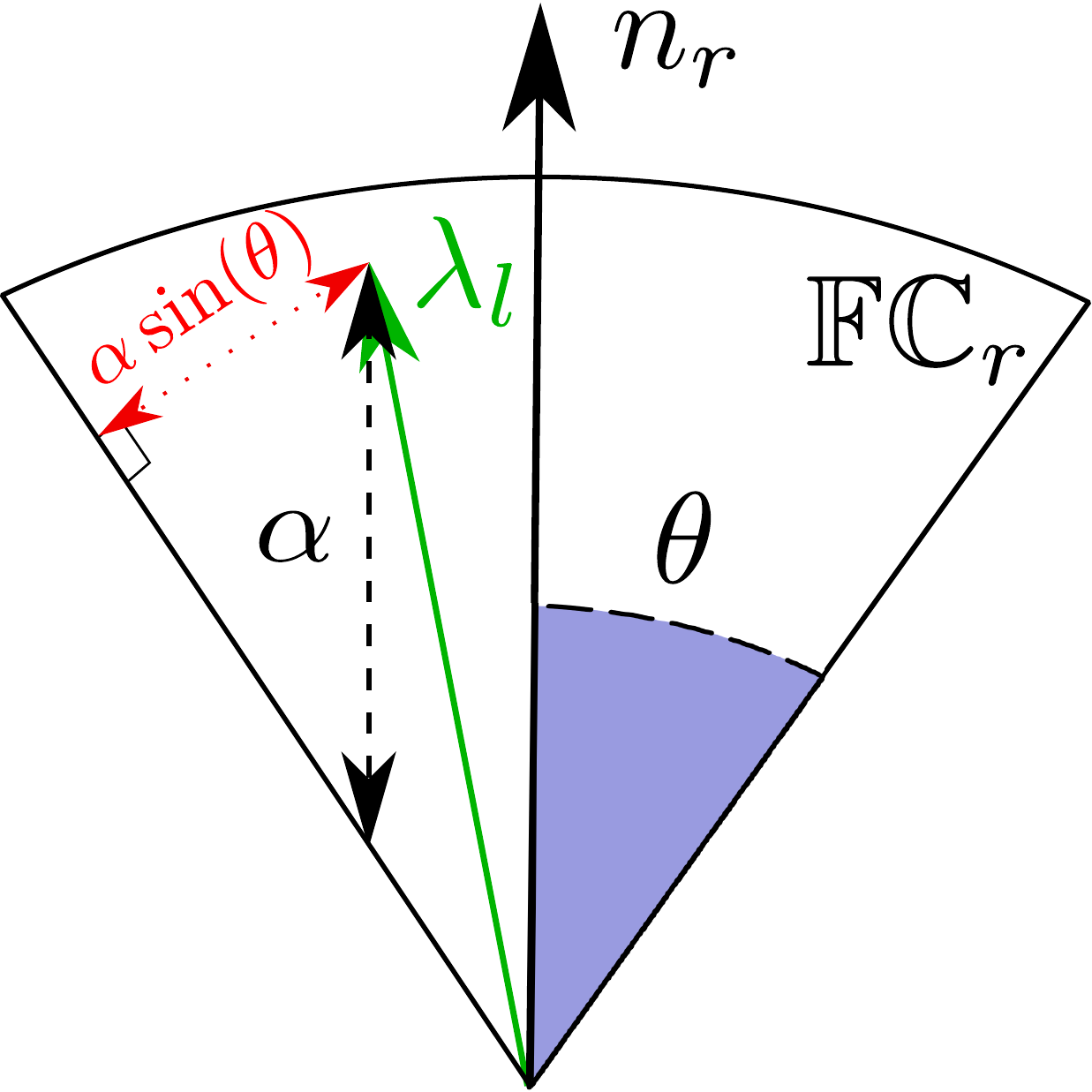}
    	\captionof{figure}{\textit{Left}: Friction polyhedron constraint. 
        \textit{Right}: Illustration of the $\alpha$ lower-bound of the 
        friction cone stability margin.}
    	\label{fig:figcone}
    	\vspace{-12pt}
  	\end{minipage}
    \label{fig:f2}
\end{figure*}

\subsubsection{Kinematic constraints}\label{sec:kin}
In order to ensure kinematic reachability, we must account for the workspace of each
independent leg. To do this, we constrain each contact location within the biggest
square inscribed in the leg workspace, as shown in \fref{fig:figkin} (\textit{right}). Algebraically:
\begin{equation}\label{eq:kin}
\left | \mathbf{f}_{i} - \left[ \mathbf{r}_{T(i)} + L_i \begin{pmatrix} 
\cos(\theta_i + \phi_i) \\
\sin(\theta_i + \phi_i)
\end{pmatrix} \right] \right| \leq d_{lim},
\end{equation}
where $\mathbf{r}_{T(i)}$ is the \gls{com} location after transitioning to the contact 
(given by the gait matrix $\mathbf{T}(i)$), $d_{lim}$ is the square diagonal, $L_i$ is the 
approximate distance from the trunk to the leg, and $\phi_i$ is a known offset for each 
foot. Here, the trigonometric functions are decomposed in terms piecewise linear
approximations of the trunk orientation functions. \vskip 6pt

To have them expressed as mixed-integer convex, we will follow the approach of 
\cite{deits2014footstep} and replace the trigonometric functions of \eqref{eq:kin} with
piecewise linear approximations of $N_s$ segments $s$ and $c$. We define binary
matrices $\mathbf{S}$ and $\mathbf{C}$ in $\{0,1\}^{N_f \times N_s}$ to assign linear
segments, which is done with the following constraints:
\begin{equation}
\sum_{s = 1}^{N_s}{\mathbf{S}_{i s}} = 1 \hspace{25pt} \sum_{s = 1}^{N_s}{\mathbf{C}_{i 
s}} = 1,
\end{equation}
\vspace{-6pt}
\begin{equation}
\mathbf{S}_{i k} \Rightarrow \begin{cases}
\psi_{k-1} \leq \theta_i \leq \psi_{k}\\
s_i = m_{s_k}\theta_i + n_{s_k}
\end{cases} \mathbf{C}_{i k} \Rightarrow \begin{cases}
\gamma_{k-1} \leq \theta_i \leq \gamma_{k}\\
c_i = m_{c_k}\theta_i + n_{c_k}
\end{cases},
\end{equation}
where $\psi$ and $\gamma$ represent the boundaries between each linear segment, and
$m$ and $n$ represent its slope and intersection.\vskip6pt

\subsection{End-effector Trajectories}\label{sec:effec}

As mentioned in \sref{sec:contact}, the end-effector trajectories are defined by the gait transition 
matrix $\mathbf{T}$. For simplicity, we define the function $\gamma(j,t)$ to reference the knots 
over which an end-effector swings between two adjacent contacts in the plan, 
where $j$ indicates the time-slot used for the swing and $t \in [1, \dots, N_k]$ 
indicates the knot, where $N_k$ is the number of knots allocated for each slot. 
Then, the end-effector motions are governed by the following 
constraint:
\begin{equation}
\mathbf{T}_{i j} \Rightarrow \mathbf{p}_{l(i)\gamma(j,N_k)} = \mathbf{f}_i,
\end{equation}
where $l(i)$ is the leg number for the $i^{th}$ contact. This constraint enforces that the 
leg reaches the contact position $\mathbf{f}_i$ at the end of the $j^{th}$ slot. Also, it is 
important to constrain that the leg remains stationary when there is no transition. This 
is enforced using the following constraint over the $l^{th}$ leg:
\begin{equation}
\sum_{i \in C(l)}^{ } \mathbf{T}_{i j} = 0 \Rightarrow \mathbf{p}_{l \gamma(j,t)} = 
\mathbf{p}_{l \gamma(j,1)} \  \forall t \in [2,\dots,N_k],
\end{equation}
where $C(l)$ are the contact indexes assigned to the $l^{th} $leg. Further constraints can 
be added to make the end-effector follow a specific swing trajectory. Moreover, we ensure 
kinematic feasibility by constraining the \gls{com} position with respect to the end-effectors 
(Fig. 6). Here, we use the bounding box constraint:
\begin{equation}
\textbf{d}_{-} < \mathbf{r}_j - \frac{\sum_{l = 1}^{n_l} \mathbf{p}_{l j}}{n_l} < 
\textbf{d}_{+},
\end{equation}
where $\textbf{d}_{-}$ and $\textbf{d}_{+}$ are the bounding box limits.

\subsection{Contact Dynamics}\label{sec:frict}

We model the dynamic interaction of the forces through activation and friction cone constraints, 
which ensure stability in the motion. Here, we will describe how to pose these 
constraints as mixed-integer convex. \vskip 6pt

\subsubsection{Activation constraints}
In order to ensure dynamic consistency, it is important that no force is present on
a leg when it breaks contacts. This is constrained, for the $l^{th}$ leg, as:
\begin{equation}\label{eq:cc}
\sum_{i \in C(l)}\mathbf{T}_{i j} = 1 \Rightarrow \boldsymbol{\lambda}_{l \gamma(j,t)} = 0 \ 
, \ \forall t \in NC(j),
\end{equation}
where $NC(j)$ is the set of knots in the $j^{th}$ slot used for the swing. 
This relation can be seen as a complementarity constraint, which can be
also used to activate contacts \cite{Posa14,Mastalli16}, between 
$\sum_{i \in C(l)}\mathbf{T}_{i j}$ and $\boldsymbol{\lambda}_{l \gamma(j,i)}$. \vskip6pt

\subsubsection{Dynamic stability constraints}
In order to maintain stability in non-coplanar contact conditions, the
contact forces must remain within their friction cones \cite{wieber2002stability}.
We exploit the convex segmentation of the terrain in order to linearly approximate
the friction cone constraints at each segment. This allows us to maintain the convexity
in the problem, while also providing formal robustness guarantees.\vskip6pt

Given that the $\mathrm{r}^{th}$ surface has a normal unit vector $\hat{\mathbf{n}}_\mathrm{r}$, 
we approximate its friction cone as a polyhedron with $N_e$ edges $\mathbf{v}_{r_1}, 
\dots, \mathbf{v}_{r_{N_e}}$. Therefore, each force remains in its respective 
contact cone $\mathbb{FC}_\mathrm{r}$ through the following constraint:
\begin{equation}\nonumber
  \boldsymbol{\lambda}_{l j} \in \mathbb{FC}_\mathrm{r} \Rightarrow  
  \boldsymbol{\lambda}_{l j} = \sum_{e = 1}^{N_e} \rho_e \mathbf{v}_{\mathrm{r}_e} \ , \ \rho_1, \dots, 
  \rho_{N_e} > 0,
\end{equation}
where $\rho_e$ are positive multipliers on each cone edge, as shown in \fref{fig:figcone} 
(\textit{left}). Then, in order to add robustness to the motion, we maximize the distance
between the nonlinear friction cone boundary and the force vector. This distance can be
computed as $\alpha\sin\theta$, where $\theta$ is the half-angle of the cone, and $\alpha$ is
defined as
\begin{equation}\nonumber
\alpha = \text{arg}\max_{\bar{\alpha}}\text{ s.t }\boldsymbol{\lambda}_{l j} - \bar{\alpha} \hat{\mathbf{n}}_\mathrm{r} \in \mathbb{FC}_\mathrm{r},
\end{equation}
namely, $\alpha$ is the distance from the contact force $\boldsymbol{\lambda}_{l j}$ to
the boundary of the friction cone, along the normal force direction \cite{del2016fast},
as shown in \fref{fig:figcone} (\textit{right}). Therefore, we seek to maximize the
value of $\alpha$ at each knot, which increases the stability margin, and we introduce
the following linear constraint over each safe surface:
\begin{equation}
\mathbf{T}_{ij} \ \text{and} \ \mathbf{H}_{ri} \Rightarrow \boldsymbol{\lambda}_{l(i) \gamma(j)}
- \alpha_{l(i) \gamma(j)} \hat{\mathbf{n}}_\mathrm{r} \in \mathbb{FC}_\mathrm{r} \ , \ \alpha \geq 0 .
\end{equation} 

Since the contact cone must not change when it is in stance phase, we also add the
constraint:
\begin{equation}\nonumber
\sum_{l \in C(i)} \sum_{t \in NS(j)} \mathbf{T}_{l t} = 0 %
,
\end{equation}
\begin{equation}
\Rightarrow \boldsymbol{\lambda}_{l(i) \gamma(j)} - \alpha_{l(i) \gamma(j)} \mathbf{n}_
\mathrm{r} \in \mathbb{FC}_\mathrm{r},
\end{equation} 
where $NS(j)$ is the set of time-slots succeeding $j$. Note that this ensures that the cone 
constraint holds for all the succeeding knots in which the leg is stationary. \vskip 6pt

\subsection{Convex Decomposition of Angular Dynamics}\label{sec:conv_ang_momentum_model}

The angular dynamics of the centroidal model are non-convex due to the cross product 
$(\mathbf{p}_{l} - \mathbf{r}) \times \boldsymbol{\lambda}_{l}$. Other mixed-integer 
approaches have used McCormick envelopes to approximate these bilinear relations 
\cite{valenzuela2016mixed}. Nevertheless, this increases significantly the complexity of 
the problem. Here, we exploit the relation noted in \cite{ponton2016convex}, namely that 
this non-convexity can be addressed by reformulating bilinear constraints as a 
decomposition of quadratic terms:
\begin{equation}\label{eq:cdc}
  \mathbf{ab} = \frac{\mathbf{u}^+ - \mathbf{u}^-}{4} \ \
  \mathbf{u}^+ \geq (\mathbf{a} + \mathbf{b})^2 \ \
  \mathbf{u}^- \geq (\mathbf{a} - \mathbf{b})^2.
\end{equation}

This convex decomposition is represented via quadratic constraints, 
turning the problem into \gls{miqcqp} and, thus, a convex optimization. We penalize the norm of $\mathbf{u}^+$ and $\mathbf{u}^-$ to bound the rate of the centroidal angular momentum. For more details, the readers can refer to \cite{ponton2016convex}. \vskip6pt

\subsection{Approximate Torque Limits}\label{sec:torq}

In order to improve the feasibility of the motion, we approximate torque limits by relying 
on a quasi-static motion assumption. This  allows us to represent torques by projected
the contact forces at each joint, compensating gravity terms, with respect to each leg
frame:
\begin{equation}\nonumber
\mathbf{J}_{l,j}^T \boldsymbol{\lambda}_{l,j} \leq \boldsymbol{\tau}_{max},
\end{equation}
where $\mathbf{J}_{l,j}\in\mathbb{R}^{3\times 3}$ is the operational space foot Jacobian
for the $l^{th}$ leg at the $j^{th}$ knot, and $\boldsymbol{\tau}_{max}$ are the
joint torque limits of the leg. Unfortunately, convex computation of the Jacobian
requires prior knowledge of the joint positions. Thankfully, the invariance of
the operational space allows us to 
approximate it around a nominal value $\mathbf{J}_{l,j}^*$. 
Therefore, we approximate joint limits with the following constraint:
\begin{equation}
{\mathbf{J}^*}_{l,j}^{T} \boldsymbol{\lambda}_{l,j} \leq \boldsymbol{\tau}_{max},
\end{equation}

This constraint approximates the \gls{awp} to define an approximation of the \gls{fws}, 
as defined in \cite{Orsolino18}. In practice, this approximation is useful for most
motions, close to the nominal position and without aggressive speeds. For further
precision, one could use robust optimization \cite{bertsimas2011theory} to constrain
over the variations of the Jacobian ${\mathbf{J}^*}_{l,j} \pm \delta {\mathbf{J}}_{l,j}$.

\subsection{Trajectory Optimization}

Given the constraints stated above, we formulate a convex trajectory optimization, as 
follows:
\begin{equation}\nonumber
\underset{\mathbf{r}, \mathbf{k}_o, \mathbf{p}_l, \boldsymbol{\lambda}_l}{\text{min}} 
\ g_T + \sum_{k = 1}^{N} g(k),
\end{equation}
where $T$ is the total number of knots, $g(k)$ is a running cost along the plan and
$g_T$ is a terminal cost. Our running cost $g(k)$ maximizes the stability of the motion,
while seeking for the fastest and smoothest gait. For this, the objectives are:
\begin{enumerate}
    \item Minimize the \gls{com} acceleration $\ddot{\textbf{r}}$.
    \item Minimize the contact forces magnitude $\|\boldsymbol{\lambda}\|$.
    \item Minimize the upper bound of quadratic terms $\textbf{U}=(\mathbf{u}^-, 
    \mathbf{u}^+)$ used in the convex angular dynamics model.
    \item Maximize the stability margin $\alpha$.
    \item Minimize the execution time, by minimizing the sum of the elements in the time 
    vector $\textbf{t}$.
\end{enumerate}
The running cost $g(k)$ is defined as:
\begin{equation}\nonumber
g(k) = \|\ddot{\mathbf{r}}_k\|_{\mathbf{Q}_v} + 
		 \|\boldsymbol{\lambda}_{l,k}\|_{\mathbf{Q}_F} +
         q_u \mathbf{U}_k + q_t t_k - q_\alpha \alpha_k,
\end{equation}

On the other hand, the terminal cost $g_T$ biases the plan towards its goal, the 
terminal position $\mathbf{r}_G$, as:
\begin{equation}
    g_T = \|\mathbf{r}_T - \mathbf{r}_G\|_{\mathbf{Q}_{g}},
\end{equation}
where $q$ are positive weights, $\mathbf{Q}$ are positive-semidefinite weighting
matrices, and $\|\mathbf{v}\|_\mathbf{Q}$ stands for the weighted squared-$L_2$ norm
$\mathbf{v}^T\mathbf{Q}\mathbf{v}$. In practice, we add a small cost to $\| \dot{k} \|_{Q_k}$
in order to generate smoother motions.

\begin{figure*}[t]
  \centering
  \includegraphics[width=0.87\textwidth]{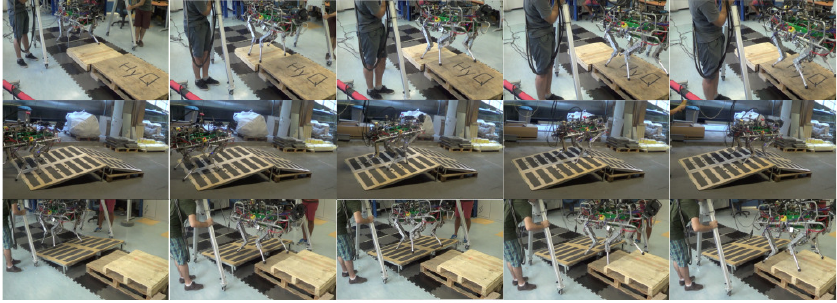}
  \caption{Snapshots of experiments 1 through 3, used to evaluate the performance of our planner
  on challenging terrain. \textit{Top}: Climbing a pallet and crossing a \unit[20]{cm} gap,
  while climbing \unit[3]{cm}. \textit{Middle}: Climbing a $10^\circ$ slope and crossing
  a \unit[15]{cm} gap leading to a $-10^\circ$ slope. \textit{Bottom}: Climbing a
  10$^\circ$ slope and crossing a \unit[20]{cm} gap, while climbing down \unit[6]{cm}
  leading to an \unit[11]{cm} stair.}
  \label{fig:figexpers}
  \vspace{-9pt}
\end{figure*}

\section{Whole-body Control}
The \gls{com} motion, body attitude and swing motions are controlled by a
\textit{trunk controller}. It computes the feed-forward joint torques
$\boldsymbol{\tau}_{ff}^*$ necessary to achieve a desired motion without
violating friction, torques or kinematic limits. To address unpredictable events
(e.g. limit foot divergence in case of slippage on an unknown surface), an
impedance controller computes in parallel the feedback joint torques
$\boldsymbol{\tau}_{fb}$ from the desired joint motion
$(\mathbf{q}^d_j, \dot{\mathbf{q}}^d_j)$. Note that the desired body and joint motions
have to be consistent with each other in order to prevent conflicts with the
trunk controller.

To achieve compliantly desired trunk motions, we compute a reference \gls{com}
acceleration ($\ddot{\mathbf{r}}^r\in\mathbb{R}^3$) and body angular
acceleration ($\dot{\boldsymbol{\omega}}^r_b\in\mathbb{R}^3$) through a
\textit{virtual model}:
\begin{align}\nonumber
	\ddot{\mathbf{r}}^r &= \ddot{\mathbf{r}}^d + 
    	\mathbf{K_r} (\mathbf{r}^d - \mathbf{r}) +
        \mathbf{D_r} (\dot{\mathbf{r}}^d - \dot{\mathbf{r}}),\\
	\dot{\boldsymbol{\omega}}^r_b &= \dot{\boldsymbol{\omega}}^d_b +
    	\mathbf{K}_{{\theta}} e(\mathbf{R}_b^d \mathbf{R}_b^T) +
		\mathbf{D}_{{\theta}} ({\omega}^d_b - \boldsymbol{\omega}_b),
\label{eq:virtualModel}
\end{align}
where ($\mathbf{r}^d,\dot{\mathbf{r}}^d,\ddot{\mathbf{r}}^d)\in\mathbb{R}^3$
are the desired \gls{com} position, velocity and acceleration respectively, 
$e(\cdot): \mathbb{R}^{3\times 3}\rightarrow \mathbb{R}^3$ is a mapping from the
rotation matrix into the associated rotation vector,
$\boldsymbol{\omega}_b\in\mathbb{R}^3$ is the angular velocity of the trunk.
$\mathbf{K_r}, \mathbf{D_r}, \mathbf{K}_{{\theta}}, 
\mathbf{D}_{{\theta}}\in\mathbb{R}^{3\times 3}$ are positive-definite
diagonal matrices of proportional and derivative gains, respectively.

The target of our trunk controller is to minimize the error between the
reference and actual accelerations while enforcing friction, torque and kinematic
constraints. As mentioned above, the reference accelerations are computed from
\eqref{eq:virtualModel}. We formulate the problem using \gls{qp} with the generalized
accelerations and the contact forces as decision variables, i.e.
$\mathbf{x}=[\ddot{\mathbf{q}}^T,\boldsymbol{\lambda}^T]^T\in\mathbb{R}^{6+n+3n_l}$:
\begin{equation}
\begin{aligned} 
	\mathbf{x}^* = \ & \text{arg}\min_{\mathbf{x}} g_{err}(\mathbf{x}) +
    	\|\mathbf{x}\|_\mathbf{W}\\
	 	\st \quad & \mathbf{Ax} = \mathbf{b} \\
				&\underline{\mathbf{d}} < \mathbf{Cx} < \bar{\mathbf{d}}
\end{aligned} 
\label{eq:qp}
\end{equation}
where $n$ represents the number of active \gls{dofs}. The first term of the cost function
\eqref{eq:qp} penalizes the \textit{tracking}
error:
\begin{equation}
g_{err}(\mathbf{x}) = 
\left\|
\begin{array}{c}
\ddot{\mathbf{r}} - \ddot{\mathbf{r}}^r \\
\dot{\boldsymbol{\omega}}_b - \dot{\boldsymbol{\omega}}_b^r
\end{array}
\right\|_\mathbf{S},
\label{eq:tracking_cost}
\end{equation} 
while the second one is a regularization factor to keep the solution bounded
or to pursue additional criteria. Both costs are quadratic-weighted terms. 
As the \gls{com} acceleration is not a decision variable, we
compute them from the contact forces using the centroidal dynamic model.
We then re-write the tracking cost \eqref{eq:tracking_cost} as
$\|\mathbf{Gx}-\mathbf{g}_0\|$ where:
\begin{align}\nonumber
\mathbf{G} &=
	\mat{\mathbf{0}_{3\times 3}	&	\mathbf{0}_{3\times 3}	&	\mathbf{0}_{3\times n}
    	&	\frac{1}{m}\mathbf{I}_1 \cdots \frac{1}{m}\mathbf{I}_{n_l} \\
		 \mathbf{0}_{3\times 3} &	\mathbf{1}_{3\times 3}	&	\mathbf{0}_{3\times n}
        &	\mathbf{0}_{3\times 3n_l}},
\mathbf{g}_0 =
	\mat{\ddot{\mathbf{r}}^r + \mathbf{g} \\  \dot{\boldsymbol{\omega}}_b^r},
\label{eq:costMatrix}
\end{align}
and $\mathbf{I}_k$ representing an identity matrix for the $k^{th}$ end-effector.
The equality constraints $\mathbf{Ax}=\mathbf{b}$ encodes dynamic consistency,
stance condition and swing task. On the other hand, the inequality constraints
$\underline{\mathbf{d}} < \mathbf{Cx} < \bar{\mathbf{d}}$ encode friction, torque,
and kinematic limits.

We map the optimal solution $\mathbf{x}^*$ into desired feed-forward joint torques
$\boldsymbol{\tau}_{ff}^*\in\mathbb{R}^n$ using the actuated part of the full dynamics
of the robot as:
\begin{equation}
	\boldsymbol{\tau}^*_{ff} = \mat{\mathbf{M}_{bj}^T & \mathbf{M}_j}\mathbf{\ddot{q}}^*
    + \mathbf{h}_j - \mathbf{J}_{cj}^T\boldsymbol{\lambda}^*
\end{equation}
where $\mathbf{M}_{bj}\in\mathbb{R}^{(6+n)\times n}$ represents the
coupled inertia between the floating-base and joints,
$\mathbf{M}_j\in\mathbb{R}^{n\times n}$ the joint contribution to the
inertia matrix, $\mathbf{h}_j\in\mathbb{R}^n$ is the force vector that
accounts for Coriolis, centrifugal, and gravitational forces to the joint torque,
and $\mathbf{J}_{cj}\in\mathbb{R}^{3n_l\times n}$ is a stack of Jacobians of
the $n_l$ end-effectors.

Finally, the feed-forward torques $\boldsymbol{\tau}_{ff}^*$ are summed with the
joint PD torques (i.e. feedback torques $\boldsymbol{\tau}_{fb}$) to form the
desired torque command $\boldsymbol{\tau}^d$:
\begin{equation}
	\boldsymbol{\tau}^d = \boldsymbol{\tau}^*_{ff} +
    PD(\mathbf{q}^d_j, \dot{\mathbf{q}}^d_j),
\end{equation}
which is sent to a low-level joint-torque controller. For more information on this controller the reader can refer to \cite{mastalli:hal-01649836}

\section{Experimental Validation}

We validated our approach on the \gls{hyq} robot \cite{semini2011design}, a
\unit[85]{kg} hydraulically actuated quadruped robot, traversing various
challenging terrains while also optimizing the gait transition \footnote{video with the experiments: https://youtu.be/s4PMXpbwUes}. 
The \gls{hyq} robot is fully-torque controlled and equipped
with high-precision joint encoders, a Multisense SL sensor (Carnegie Robotics)
and an Inertial Measurement Unit (KVH). \gls{hyq} has approximately the
dimensions of a goat: \unit[1.0]{m}$\times$\unit[0.5]{m}$\times$\unit[0.98]{m} 
(length$\times$width$\times$height). The planned motions are computed off-board
using the Drake Toolbox \cite{drake} in MATLAB 2015a, on a Dual-Core
Laptop running Mac OS X Sierra. We use Gurobi 6.0.5 \cite{gurobi} as our
\gls{miqcqp} solver. The whole-body controller and state estimation run fully on-board and in
real-time. Our state estimation is based on a modular inertial-driven Extended Kalman
Filter which incorporates a rugged probabilistic leg odometry component with additional
inputs from stereo vision and LIDAR registration, as described in \cite{2017RSS_nobili}.\vskip6pt

\subsection{Locomotion on rough terrain}
In our first experiments we showcase the different capabilities of our approach
by navigating over various challenging terrains composed of gaps, ramps, and steps with different
terrain heights (\fref{fig:figexpers}). Our method allows the robot to successfully cross
those terrains because of the use of a) a convex model of robot's dynamics and terrain, b) an
approximation of the torque limits, and c) stability maintenance capabilities in
non-coplanar contact conditions.\vskip6pt

\subsubsection{Convex model of robot's dynamics and terrain}
Traversing a gap while climbing requires to accommodate properly the body
position because of the robot's kinematic limitations. In \fref{fig:figexpers}
(\textit{top}), we show that our method exploits properly the leg reachability,
requiring fewer contact locations to reach the goal compared with
\cite{mastalli2017trajectory}. In fact, having a convex model of the robot's dynamics,
and especially of the terrain, avoids to fall into a local minimum because of the global
optimality guarantees. This convex terrain model works well for most
scenarios, however, it might lose validity when the terrain has significant 
non-linear curvature.\vskip6pt

\begin{figure}[t]
  \vspace{-6pt}
  \centering
  \includegraphics[width=0.99\linewidth]{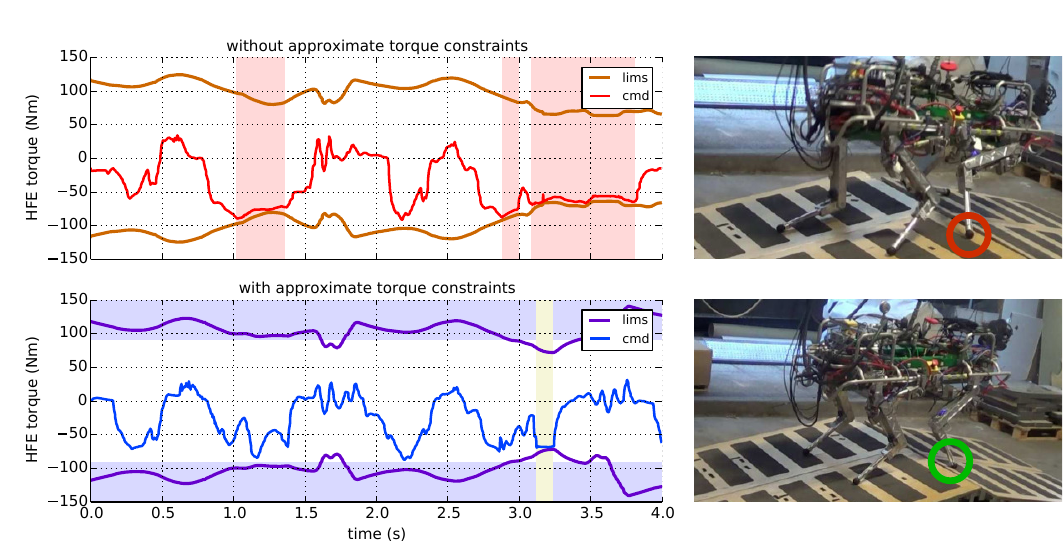}
  \caption{Comparison of climbing up and down two slopes with (\textit{bottom})
  and without (\textit{top}) approximate torque constraints. Figures show the
  real torque limits (\textit{lims}) and the torque (\textit{cmd}) 
  for the RF hip pitch joint (HFE).}
  \label{fig:figtl}
  \vspace{-18pt}
\end{figure}

\subsubsection{Approximate torque constraints}
We compare two different planned motions, with and without approximate
torque constraints in the optimization (see \fref{fig:figtl}). While the robot climbed up
successfully the slope in both cases, it was able to cross the gap only when the torque
limits were considered. In fact, in these scenarios, the admissible set of contact forces
is reduced due to the friction cones pointing outwards from the robot's \gls{com}.
Here, only the approximate torque constraints 
(highlighted in blue in \fref{fig:figtl}) account for this reduction of the 
feasible wrench set, and as result, the contact forces are distributed accordingly, see
\fref{fig:figtl} (\textit{bottom})\footnote{Note that \gls{hyq}'s torque limits depend on
the joint position, resulting in limit differences in the
yellow shade on \fref{fig:figtl} (\textit{bottom}), 
which are not accessible using the centroidal dynamics and the quasi-static assumption.}.  
However, if we do not consider the approximate torque
constraints, the robot reaches torque limits three times, 
red shade in \fref{fig:figtl} (\textit{top}), and it falls while 
climbing down, see \fref{fig:figtl} (\textit{top-right}). \vskip6pt

\begin{figure*}[t]
  \centering
  \includegraphics[width=0.85\textwidth]{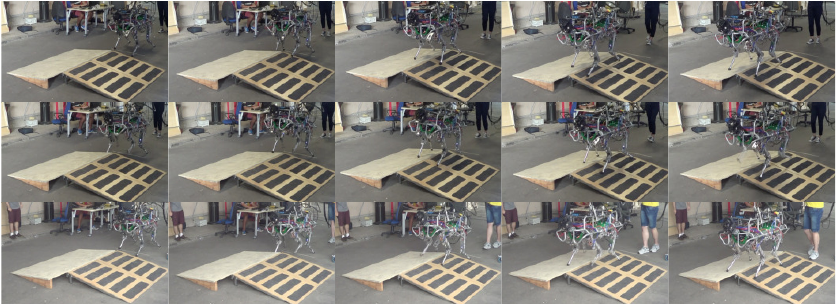}
  \caption{Snapshots of experiment 4, used to evaluate the 
  performance of different gaits generated with our planning scheme in a
  roof-like terrain. \textit{Top}: Fixed walking gait. \textit{Middle}:
  Fixed trotting gait. \textit{Bottom}: Automatically discovered gait.}
  \label{fig:figexpgait}
  \vspace{-9pt}
\end{figure*}

\subsubsection{Stability in non-coplanar contact conditions}
Crossing a gap and ascending up stairs is a challenging task, because
the robot has to maintain stability while considering potential
non-coplanar contacts, see \fref{fig:figexpers} (\textit{bottom}). 
For instance, \gls{zmp}-based approaches such as
\cite{mastalli2017trajectory,winkler17b} cannot ensure stability in
such terrains. We experimentally found an increment on the
robustness of the planned motion compared to the aforementioned methods. 
This robustness is especially important because
we rely on a reference for the robot attitude modulation as in
\cite{mastalli2017trajectory}. This is done because we restrict our trajectories to
small angular momentum changes, because of
limitations in the convex model (see \sref{sec:conv_ang_momentum_model}).

\subsection{Gait transitions}

Afterwards, we tested the gait planning and transitioning capabilities
of our approach. In the optimization, 
we seed a gait sequence by first optimizing only over the linear dynamics. 
For this, we run trials over a challenging terrain with 
non-coplanar contacts to compare two different scenarios: a) crossing with a fixed 
gait that is predefined, walk and trot, and b) crossing while the gait sequence is optimized. 
Figure \ref{fig:figexpgait} shows snapshots of these experimental trials.\vskip4pt

\subsubsection{Fixed walking and trotting gaits}
In these cases, as expected, the motions are executed robustly across
the entire trial. For walking, we obtained motions of quality similar to those
of the previous experiments. In the case of the trotting, we increased the
execution speed compared to the walking case (from \unit[18]{cm/s} to 
\unit[31]{cm/s}). This fast motion is effectively tracked by our controller, 
despite relatively slow corrections, from the visual sources, in the state
estimation. \vskip4pt

\subsubsection{Automatic gait discovery}
\begin{figure}[t]
	\centering
	\includegraphics[trim={50pt 0 50pt 0},clip,width=0.99\linewidth]{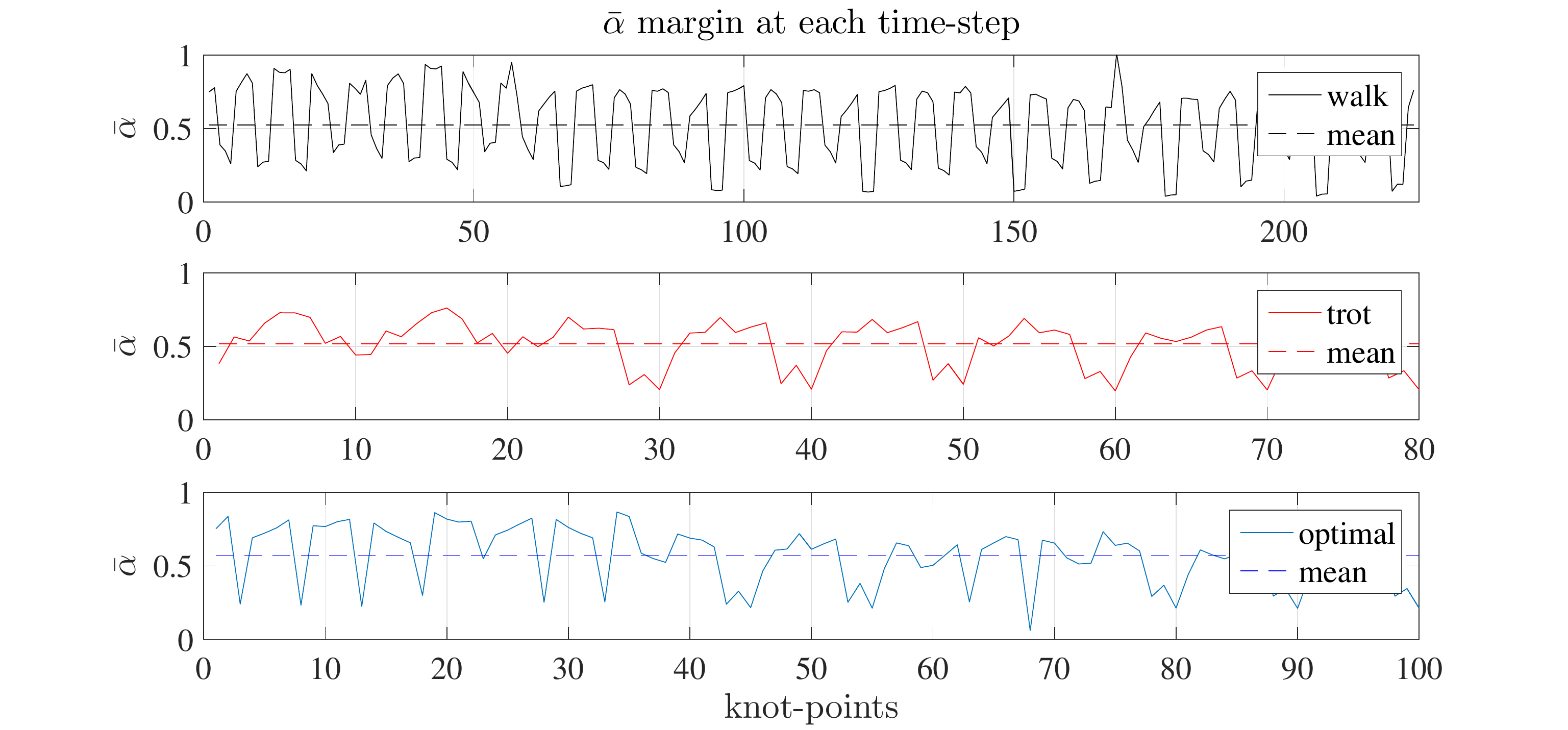} 
    
    \vspace{6pt}
    
    \includegraphics[width=0.95\linewidth]{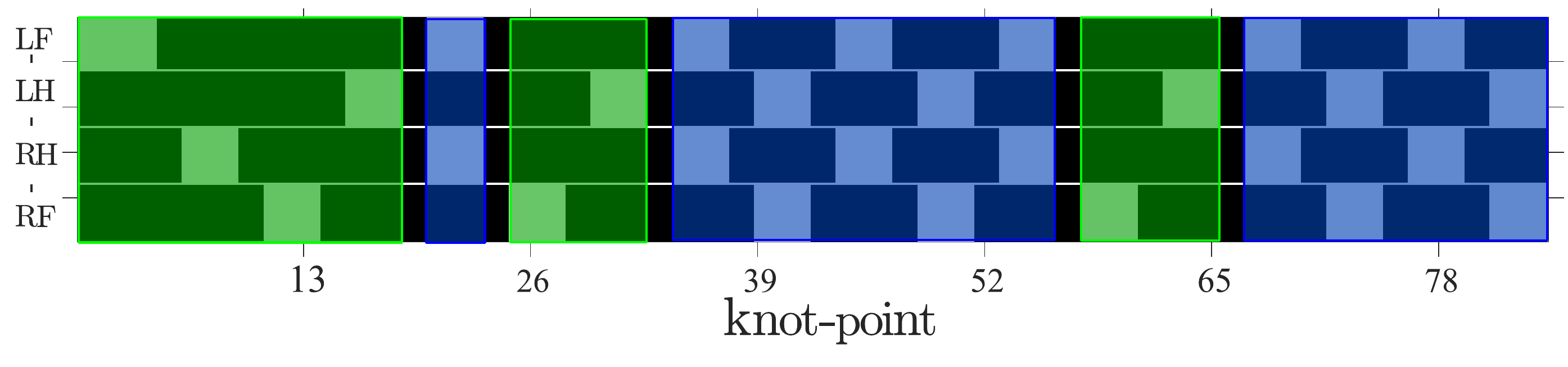}
    \caption{\textit{Top:} Normalized $\alpha$ margin for 
    different gaits. \textit{Bottom:} Resulting optimal gait 
    sequence for navigating in roof-like terrain, blue and green 
    segments represent trot and walk, respectively.}
    \label{fig:figgaitplan}
    \vspace{-12pt}
\end{figure}

In this trial, we showcase the trade-offs involved when optimizing the gait
transitions \fref{fig:figexpgait} (\textit{bottom}). First, in order to minimize
acceleration, our motion planner selects a walking gait before smoothly
transitioning to a dynamic trotting gait. As expected, the robot continues trotting
until it faces a significant change of terrain. At that moment, the robot 
transitions to a slower gait (i.e. walking gait) in order to accommodate for
the terrain conditions. For comparison, \fref{fig:figgaitplan} shows the minimum $\alpha$
margin at each time-step, normalized by its contact force. 
Finally, it returns to a trotting gait as is detailed in
\fref{fig:figgaitplan}. 
In contrast to \cite{aceituno2017convex}, we do not include
any terrain heuristics in the planning stage, resulting in an automatic gait discovery
based on the centroidal dynamics of the robot. 

\subsection{Discussion}

The experimental trials presented above showcase the main aspects of our approach. 
As expected, all of our plans result in successful executions. Thanks to our incorporation
of approximate torque limits, we are able to push the limits of our hardware, 
reaching faster walking speeds (around \unit[15]{cm/s}) than previous approaches executed 
on the same robot  \cite{winkler2015planning,mastalli2017trajectory,winkler17b}. 
Additionally, we have shown that our approach is able to handle gait transitions
robustly even on challenging terrain. In our experiments, the convex approximation of the
angular dynamics helps the planner to generate natural motions. Nonetheless, we found that
this model does not extend well to more dynamical gaits, such as bounding or pace,
since it cannot impose limits in the angular momentum, causing divergence.

\begin{table}[t]
	\centering
    \caption{Computation time for a locomotion cycle for various terrain
    conditions and gaits}
    \begin{tabular}{| c || c c c |}
    \hline
    \textbf{Experiment} & convex surfaces & Gait & \textbf{mean time (s)} \\
    \hline
    \hline
    Exp. 1 & 3 & Walk & 0.47 \\
    Exp. 2 & 3 & Walk & 0.64 \\
    Exp. 3 & 4 & Walk & 0.44 \\
    \hline
    \hline
    Exp. 4 & 3 & Walk & 0.48 \\
    Exp. 4 & 3 & Trot & 0.51 \\
    Exp. 4 & 3 & Free & 1.62 \\
    \hline
    \end{tabular}
    \label{table:time}
    \vspace{-15pt}
\end{table}

Computation time is one of the main concerns in any optimization problem. 
In \tref{table:time}, we report the required computation time (for a single
locomotion cycle one stride) of the above scenarios. We plan motion at least two orders
of magnitude faster than previous work on similar terrains
\cite{mastalli2017trajectory,neunert2017trajectory} because of our convex
formulation using \gls{miqcqp}. The kinematic constraints imposed help the
solver to quickly discard infeasible surfaces. 
This captures the main advantage of our method, that it is able to efficiently 
plan motions through 3D environments with obstacles, while maintaining 
stability despite changes in the terrain.

Other factors might influence the computation time. For instance, leaving the
gait as a decision variable increases computation time significantly. 
As in \cite{deits2014footstep}, we found that the incorporation of
piecewise-affine trigonometric functions has a negative effect on the
computation time. Likewise, quadratic constraints in \eqref{eq:cdc} also influence the
computation time, resulting in increases of around 500\%. However, we believe
that our method can potentially be extended to receding horizon fashion.\vskip 6pt

\section{Conclusions}

We have presented a novel approach for simultaneously planning contacts
and motions on multi-legged robots based on \gls{micp}. Our approach is able
to handle complex terrain, while also providing formal robustness guarantees
on the plan. Moreover, we incorporate the gait sequences
as a decision variable, which allows for automatic gait discovery. 
Our experimental trials use both a state-of-the-art whole-body controller
and state estimation \cite{mastalli:hal-01649836,2017RSS_nobili}.
We demonstrate the capabilities of our approach by traversing challenging terrains
with the \gls{hyq} robot. To our knowledge, these hardware experiments
constitute the first experimental validation of non-gaited uneven locomotion via 
friction cone stability margin. Moreover, our implementation is able to plan 
locomotion cycles in less than a second, even in complex scenarios, 
which is at least two orders of magnitude faster than previous non-convex 
approaches \cite{mastalli2017trajectory,dai2014whole} in similar environments. \vskip 6pt

In future work, we plan to expand on the capabilities of real-time planning that
this approach might offer, possibly reaching similar computation time to
\cite{winkler17b}. To this end, friction cone constraints could be relaxed to
vertex-based \gls{zmp} constraints \cite{winkler17b}, sacrificing stability
guarantees but improving computation time. Here, we have relied on
a convex decomposition of the angular dynamics. This approach works well for
simple motions where angular momentum rates are often low. However, it can
cause the angular momentum to diverge when the contact forces vary significantly.
Because of this, we are interested in exploring other models of angular dynamics,
like those proposed in \cite{valenzuela2016mixed}, so as to generate more 
aggressive behaviors. In this work, we have approximated torque limits with a nominal 
operational space Jacobian that is updated iteratively. While this works well
for the experiments performed, it loses validity for more aggressive motions.
For this reason, we are interested in studying other approximations of torque
limits. We are further interested in exploiting the piecewise affine structure
of the multi-contact dynamics in the receding horizon.

\bibliographystyle{IEEEtran}
\bibliography{IEEEabrv,references}

\end{document}